\documentclass{article}
\usepackage{custom_conference,times}
\usepackage{url}
\usepackage{amsmath}
\usepackage{amssymb}
\usepackage{graphicx}
\usepackage{subcaption}
\usepackage[utf8]{inputenc}

\usepackage[pagebackref,breaklinks=true]{hyperref} 
\renewcommand*\backref[1]{\ifx#1\relax \else (\^ #1) \fi}

\usepackage{soul} 
\usepackage[normalem]{ulem} 

\title{Clustering with Deep Learning: \\ Taxonomy and New Methods}
\author{Elie Aljalbout, Vladimir Golkov, Yawar Siddiqui, Maximilian Strobel \& Daniel Cremers\\
Computer Vision Group\\
Technical University of Munich\\
\texttt{\{firstname.lastname, cremers\}@tum.de}
}
\begin{document}
\maketitle

\begin{abstract}
Clustering methods based on deep neural networks have proven promising for clustering real-world data because of their high representational power. In this paper we propose a systematic taxonomy of clustering methods that utilize deep neural networks. We base our taxonomy on a comprehensive review of recent work and validate the taxonomy in a case study. In this case study, we show that the taxonomy enables researchers and practitioners to systematically create new clustering methods by selectively recombining and replacing distinct aspects of previous methods with the goal of overcoming their individual limitations. The experimental evaluation confirms this and shows that the  method created for the case study achieves state-of-the-art clustering quality and surpasses it in some cases.
\end{abstract}

\section{Introduction}

The main objective of clustering is to separate data into groups of similar data points. Having a good separation of data points into clusters is fundamental for many applications in data analysis and data visualization.

The performance of current clustering methods is however highly dependent on the input data. Different datasets usually require different similarity measures and separation techniques. As a result, dimensionality reduction and representation learning have been extensively used alongside clustering in order to map the input data into a feature space where separation is easier. By utilizing deep neural networks (DNNs), it is possible to learn non-linear mappings that allow transforming data into more clustering-friendly representations without manual feature extraction/selection.

In the past, feature extraction and clustering were applied sequentially \cite{ding2004k,trigeorgis2014deep}. However, recent work shows that jointly optimizing for both can yield better results \cite{song2013auto,xie2016unsupervised,yang2016towards,yang2016joint,li2017discriminatively}.

The main contribution of this paper is the formulation of a taxonomy for clustering methods that rely on a deep neural network for representation learning. The proposed taxonomy enables researchers to create new methods in a structured and analytical way by selectively recombining or replacing distinct aspects of existing methods to improve their performance or mitigate limitations. The taxonomy is in particular also valuable for practitioners who want to create a method from existing building blocks that suits their task at hand. To illustrate the value of the proposed taxonomy, we conducted a case study in which we fuse a new method based on insights from the taxonomy.

\begin{figure*}
  \includegraphics[width=\linewidth]{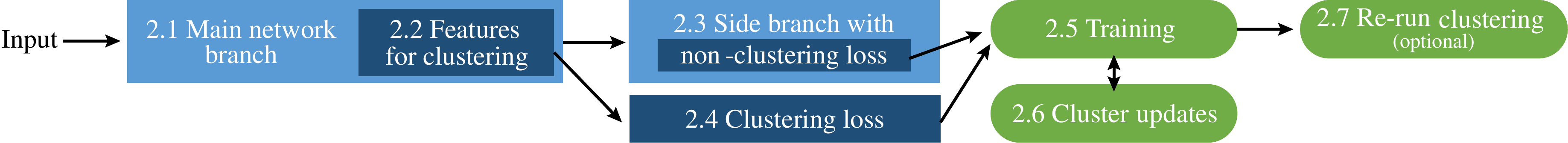}
  \caption{The general pipeline of most deep-learning-based clustering methods accodring to the proposed taxonomy, with building blocks as described in Sections~\ref{sec:arch}--\ref{sec:aftertrain}.}
  \label{fig:taxonomy}
\end{figure*}

In this case study, we use a fully convolutional autoencoder to learn clustering-friendly representations of the data by optimizing it with a two-phased training procedure. In the first phase, the autoencoder is trained with the standard mean squared error reconstruction loss. In the second phase, the autoencoder is then fine-tuned with a combined loss function consisting of the autoencoder reconstruction loss and a clustering-specific loss.

The rest of this paper is organized as follows: At first, we introduce the taxonomy and its building blocks in Section~\ref{sec:taxonomy}. In Section~\ref{sec:related}, we then present a comprehensive review of clustering methods and analyze them with respect to the proposed taxonomy. Subsequently, in Section~\ref{sec:proposed}, we propose a new method based on insights gained in a systematic way from the taxonomy. The evaluation of the proposed method is presented in Section~\ref{sec:results}, followed by the conclusions in Section~\ref{sec:conclusion}.

\section{Taxonomy}
\label{sec:taxonomy}

The most successful methods for clustering with deep neural networks all work following the same principle: representation learning using DNNs and using these representations as input for a specific clustering method. As depicted in Figure \ref{fig:taxonomy}, every method consists of the following parts, for each of which there are several options to choose from:
\begin{itemize}
    \item Neural network training procedure, consisting of:
    \begin{itemize}
        \item Main neural network branch and its usage
        \begin{itemize}
            \item Architecture of main neural network branch, described in Section~\ref{sec:arch}
            \item Set of deep features used for clustering, described in Section~\ref{sec:layer}
        \end{itemize}
        \item Neural network losses:
        \begin{itemize}
            \item Non-clustering loss, described in Section~\ref{sec:ncloss}
            \item Clustering loss, described in Section~\ref{sec:closs}
            \item Method to combine the two losses, described in Section~\ref{sec:comb}
        \end{itemize}
        \item Cluster updates, described in Section~\ref{sec:cupdates}
       
    \end{itemize}
    \item (Optional) Re-run clustering after network training, described in Section~\ref{sec:aftertrain}
    
\end{itemize}

\subsection{Architecture of Main Neural Network Branch}
\label{sec:arch}

In most DNNs-based clustering methods, the ``main branch'' of the neural network (apart from side branches towards non-clustering losses, see Section~\ref{sec:ncloss}) is used to transform the inputs into a latent representation that is used for clustering.
The following neural network architectures have previously been used for this purpose: 

\begin{itemize}
    \item \textbf{Multilayer Perceptron (MLP)}: Feedforward network, consisting of several layers of neurons, such that the output of every hidden layer is the input to next one.
    \item \textbf{Convolutional Neural Network (CNN)}:  Inspired by biology, more precisely by the organization of the animal visual cortex. Useful for applications to regular-grid data such as images, if locality and shift-equivariance/invariance of feature extraction is desired.
    \item \textbf{Deep Belief Network (DBN)}: Generative graphical model, consisting of several layers of latent variables. It is composed of several shallow networks such as restricted Boltzmann machines, such that the hidden layer of each sub-network serves as the visible layer of the next sub-network.
    \item \textbf{Generative Adversarial Network (GAN)}: A system of two competing neural network models G and D that engage in a zero-sum game. The generator G learns a distribution of interest to produce samples. The discriminator D learns to distinguish between real samples and generated ones \citep{goodfellow2014generative}.
    \item \textbf{Variational Autoencoder (VAE)}: A Bayesian network with an autoencoder architecture that learns the data distribution (generative model).
\end{itemize}

\subsection{Set of Deep Features Used for Clustering}
\label{sec:layer}

After transforming the input to a more clustering-friendly representation, the features that are then used for clustering can be taken from one or more layers of the deep neural network:
\begin{itemize}
    \item \textbf{One layer}: Refers to the case where only one layer of the network is used which is beneficial because of its low dimensionality. In most cases the output of the last layer is used.
    \item \textbf{Several layers}: Refers to the case where the representation is a combination of the outputs of several layers. Thus, the representation is richer and allows the embedded space to represent more complex semantic representations, which might enhance the separation process and help in the similarity computation \citep{saito2017neural}. 
\end{itemize}

\subsection{Non-Clustering Loss}
\label{sec:ncloss}
The non-clustering loss is independent of the clustering algorithm and usually enforces a desired constraint on the learned model. Possible options are as follows:
\begin{itemize}
    \item \textbf{No non-clustering loss}: No additional non-clustering loss function is used and the network model is only constrained by the clustering loss. For most clustering losses, the absence of a non-clustering loss can have a danger of worse representations/results, or theoretically even collapsing clusters \citep{yang2016towards}, but the latter rarely occurs in practice.
    \item \textbf{Autoencoder reconstruction loss}:  The autoencoder consists of two parts: an encoder and a decoder. The encoder maps its input $x$ to a representation $z$ in a latent space~$Z$. During training, the decoder tries to reconstruct $x$ from $z$, making sure that useful information has not been lost by the encoding phase. In the context of clustering methods, once the training is done the decoder part is no longer used, and the encoder is left for mapping its input to the latent space $Z$. By applying this procedure, autoencoders can successfully learn useful representations in the cases where the output's dimensionality is different from the input's or when random noise is injected to the input \citep{vincent2010stacked}. Additionally, they can also be used for dimensionality reduction goals \citep{hinton2006reducing}. Generally the reconstruction loss is a distance measure $d_\mathrm{AE}(x_i,f(x_i))$ between the  input~$x_i$ to the autoencoder and the corresponding reconstruction~$f(x_i)$. One particular formulation of it is using the mean squared error of the two variables:
    \begin{equation}
        L=d_\mathrm{AE}(x_i,f(x_i))=\sum_i\lVert x_i-f(x_i)\rVert^2 ,
    \end{equation}
     where $x_i$ is the input and $f(x_i)$ is the autoencoder reconstruction. This loss function guarantees that the learned representation preserves important information from the initial one, which is why reconstruction is possible.
    \item \textbf{Self-Augmentation Loss} \cite{hu2017learning} proposes a loss term that pushes together the representation of the original sample and their augmentations: 
    \begin{equation}
        L=-\dfrac{1}{N}\sum_N s(f(x),f(T(x))) ,
    \end{equation}
    where $x$ is the original sample, $T$ is the augmentation function, $f(x)$ is the representation generated by the model, and $s$ is some measure of similarity (for example cross-entropy if $f$ has a softmax nonlinearity).
    \item \textbf{Other tasks}: Additional information about training samples that is available in the form of targets, even if not perfectly suitable to dictate clustering, can be used in a (multi-task) non-clustering loss to encourage meaningful feature extraction.
\end{itemize}

\subsection{Clustering Loss}
\label{sec:closs}

The second type of functions is specific to the clustering method and the clustering-friendliness of the learned representations, therefore such functions are called clustering loss functions. The following are options for clustering loss functions:
\begin{itemize}
    \item \textbf{No clustering loss}: Even if a neural network has only non-clustering losses (Section~\ref{sec:ncloss}), the features it extracts can be used for clustering after training (Sections~\mbox{\ref{sec:cupdates}--\ref{sec:aftertrain}}). The neural network serves in this case for changing the representation of the input, for instance changing its dimensionality. Such a transformation could be beneficial for the clustering sometimes, but using a clustering loss usually yields better results \citep{xie2016unsupervised,yang2016towards}.
   
    \item \textbf{$\textbf{k}$\nobreakdash-Means loss}: Assures that the new representation is $k$\nobreakdash-means-friendly \citep{yang2016towards}, i.e.~data points are evenly distributed around the cluster centers. In order to obtain such a distribution a neural network  is trained with the following loss function:
\begin{equation}
    L(\theta)=\sum\limits_{i=1}^N \sum\limits_{k=1}^K s_{ik}\lVert z_i-\mu_k \rVert^2 ,
\end{equation}
where $z_i$ is an embedded data point, $\mu_k$ is a cluster center and $s_{ik}$ is a boolean variable for assigning $z_i$ with $\mu_k$. Minimizing this loss with respect to the network parameters assures that the distance between each data point and its assigned cluster center is small. Having that, applying $k$\nobreakdash-means would result in better clustering quality.
    
    \item \textbf{Cluster assignment hardening}:
    
    Requires using soft assignments of data points to clusters. For instance, Student's $t$\nobreakdash-distribution can be used as the kernel to measure the similarity \citep{maaten2008visualizing} between points and centroids. This distribution $Q$ is formulated as follows:
    \begin{equation}
    \label{eq:estimate}
    q_{ij}=\frac{(1+\lVert z_i-\mu_j \rVert^2/\nu)^{-\frac{\nu+1}{2}}}{\sum\limits_{j'}(1+\lVert z_i-\mu_{j'} \rVert^2/\nu)^{-\frac{\nu+1}{2}}} ,
    \end{equation}
    where $z_i$ is an embedded data point, $\mu_j$ is the $j^{th}$ cluster centroid, and $\nu$ is a constant, e.g.~$\nu=1$. These normalized similarities between points and centroids can be considered as soft cluster assignments. The \emph{cluster assignment hardening} loss then enforces making these soft assignment probabilities stricter. It does so by letting cluster assignment probability distribution $Q$ approach an auxiliary (target) distribution $P$ which guarantees this constraint. \cite{xie2016unsupervised} propose the following auxiliary distribution:
    \begin{equation}
    p_{ij}=\frac{q_{ij}^2/\Sigma_iq_{ij}}{\Sigma_{j'}(q_{ij'}^2/\Sigma_iq_{ij'})} .
    \end{equation}
    By squaring the original distribution and then normalizing it, the auxiliary distribution $P$ forces assignments to have stricter probabilities (closer to $0$ and $1$). It aims to improve cluster purity, put emphasis on data points assigned with high confidence and to prevent large clusters from distorting the hidden feature space \citep{xie2016unsupervised}. One way to formulate the divergence between the two probability distributions is using the  Kullback--Leibler divergence \citep{kullback1951information}. It is formulated as follows:
    \begin{equation}
    L=\mathrm{KL}(P\|Q)=\sum_{i}\sum_{j}p_{ij}\log\frac{p_{ij}}{q_{ij}} ,
    \end{equation}
    which is minimized for the aforementioned~$Q$ and~$P$ via neural network training.

    \item \textbf{Balanced assignments loss}: This loss has been used alongside other losses such as the previous one \citep{dizaji2017deep}. Its goal is to enforce having balanced cluster assignments. It is formulated as follows:
    \begin{equation}
    \label{eq:klu}
        L_{ba}=\mathrm{KL}(G\|U)
    \end{equation}
    where $U$ is the uniform distribution and $G$ is the probability distribution of assigning a point to each cluster:
    \begin{equation}
        g_k=P(y=k)=\frac{1}{N}\sum_iq_{ik}
    \end{equation}
    By minimizing equation \ref{eq:klu}, the probability of assigning each data point to a certain cluster is uniform across all possible clusters \citep{dizaji2017deep}. It is important to note that this property (uniform assignment) is not always desired. Thus, in case any prior is known it is still possible to replace the uniform distribution by the known prior one.
    
    \item \textbf{Locality-preserving loss}: This loss aims to preserve the locality of the clusters by pushing nearby data points together \citep{huang2014deep}. Mathematically, it is formulated as follows:
    \begin{equation}
        L_{lp}=\sum_i\sum_{j \in N_k(i)}s(x_i,x_j)\lVert z_i-z_j \rVert^2
    \end{equation}
    where $N_k(i)$ is the set of $k$ nearest neighbors of the data point $x_i$, and $s(x_i,x_j)$ is a similarity measure between the points $x_{i}$ and $x_{j}$.
    
     \item \textbf{Group sparsity loss}: It is inspired by spectral clustering  where block diagonal similarity matrix is exploited for representation learning \citep{ng2002spectral}.  Group sparsity is itself an effective feature selection method. In \cite{huang2014deep}, the hidden units were divided into $G$ groups, where $G$ is the assumed number of clusters. When given a data point $x_i$ the obtained representation has the form $\{ \phi^g(x_i)\}_{g=1}^G$. Thus the loss can be defined as follows:
     
     \begin{equation}
         L_{gs}=\sum_{i=1}^N \sum_{g=1}^G \lambda_g \lVert \phi^g(x_i) \rVert,
     \end{equation}
     where $\{ \lambda_g \}_{g=1}^G $ are the weights to sparsity groups, defined as
     \begin{equation}
         \lambda_g=\lambda \sqrt{n_g},
     \end{equation}
     where $n_g$ is the group size and $\lambda$ is a constant.
     
    \item \textbf{Cluster classification loss}: Cluster assignments obtained during cluster updates (Section~\ref{sec:cupdates}) can be used as ``mock'' class labels for a classification loss in an additional network branch, in order to encourage meaningful feature extraction in all network layers~\citep{hsu2017cnn}. 

    \item \textbf{Agglomerative clustering loss}: Agglomerative clustering merges two clusters with maximum affinity (or similarity) in each step until some stopping criterion is fulfilled.
    A~neural network loss inspired by agglomerative clustering \citep{yang2016joint} is computed in several steps. First, the cluster update step (Section~\ref{sec:cupdates}) merges several pairs of clusters by selecting the pairs with the best affinity (some predefined measure of similarity between clusters). Then network training retrospectively even further optimizes the affinity of the already merged clusters (it can do so because the affinity is measured in the latent space to which the network maps). After the next cluster update step, the network training switches to retrospectively optimizing the affinity of the newest set of newly merged cluster pairs. In this way, cluster merging and retrospective latent space adjustments go hand in hand. Optimizing the network parameters with this loss function would result in a clustering space more suitable for (agglomerative) clustering. 

\end{itemize}

\subsection{Method to Combine the Losses}
\label{sec:comb}
In the case where a clustering and a non-clustering loss function are used, they are combined as follows:
\begin{equation}
    L(\theta)=\alpha L_c(\theta)+(1-\alpha)L_n(\theta) ,
\end{equation}
where $L_c(\theta)$ is the clustering loss, $L_n(\theta)$ is the non-clustering loss, and $\alpha\in [0;1]$ is a constant specifying the weighting between both functions. It is an additional hyperparameter for the network training. It can also be changed during training following some schedule. The following are methods to assign and schedule the values of~$\alpha$: 
\begin{itemize}
    \item \textbf{Pre-training, fine-tuning}: First, $\alpha$ is set to $0$, i.e.~the network is trained using the non-clustering loss only. Subsequently, $\alpha$ is set to $1$, i.e.~the non-clustering network branches (e.g.~autoencoder's decoder) are removed and the clustering loss is used to train (fine-tune) the obtained network. The constraint forced by the reconstruction loss could be lost after training the network long enough for clustering only. In some cases, losing such constraints may lead to worse results (see Table~\ref{comparing}).
    \item \textbf{Joint training}: $0<\alpha<1$, for example $\alpha=0.5$, i.e.~the network training is affected by both loss functions.
    \item \textbf{Variable schedule}: $\alpha$ is varied during the training dependent on a chosen schedule. For instance, start with a low value for $\alpha$ and gradually increase it in every phase of the training.

\end{itemize}
In phases with $\alpha=1$, no non-clustering loss is imposed, with potential disadvantages (see \emph{No non-clustering loss} in Section~\ref{sec:ncloss}).
Similarly, in phases with $\alpha=0$, no clustering loss is imposed, with potential disadvantages (see \emph{No clustering loss} in Section~\ref{sec:closs}).

\subsection{Cluster Updates}
\label{sec:cupdates}

Clustering methods can be broadly categorized into hierarchical and partitional (centroid-based) approaches \citep{jain1999data}. Hierarchical clustering combines methods which aim to build a hierarchy of clusters and data points. On the other hand, partitional (centroid-based) clustering groups methods which create cluster centers and use metric relations to assign each of the data points into the cluster with the most similar center. 

In the context of deep learning for clustering, the two most dominant methods of each of these categories have been used. \textbf{Agglomerative clustering}, which is a hierarchical clustering method, has been used with deep learning \citep{yang2016joint}. The algorithm has been briefly discussed in Section \ref{sec:closs}. In addition, \textbf{$\textbf{k}$\nobreakdash-means}, which falls into the category of centroid-based clustering, was extensively used \citep{xie2016unsupervised,yang2016towards,li2017discriminatively,hsu2017cnn}.

During the network training, cluster assignments and centers (if a centroid-based method is used) are updated. Updating cluster assignments can have one of the two following forms:
\begin{itemize}
    \item \textbf{Jointly updated with the network model}: Cluster assignments are formulated as probabilities, therefore have continuous values between $0$ and $1$. In this case, they can be included as parameters of the network and optimized via back-propagation.
    \item \textbf {Alternatingly updated with the network model}: Clustering assignments are strict and updated in a different step than the one where the network model is updated. In this case, several scenarios are possible, dependent on two main factors:
    \begin{itemize}
        \item \textbf{Number of iterations}: Number of iterations of the chosen clustering algorithm, that are executed at every cluster update step. For instance, in \cite{xie2016unsupervised}, at each cluster update step, the algorithm runs until a fixed percentage of points change assignments between two consecutive iterations.
        \item \textbf{Frequency of updates}: How often are cluster updates started. For instance in \cite{yang2016joint}, for every $P$ network model update steps, one cluster updates step happens.
    \end{itemize}
    
\end{itemize}

\subsection{After Network Training}
\label{sec:aftertrain}

After the training is finished, even if a clustering result was produced in the process, it can make sense to re-run a clustering algorithm from scratch using the learned features, for one of several reasons:

\begin{itemize}
    \item \textbf{Clustering a similar dataset}: The general and the most trivial case is to reuse the learned features representation mapping on another dataset which is similar to the one that has been used but has different data.
    \item \textbf{Obtaining better results}: Under certain circumstances, it is possible that the results of clustering after the training are better than the ones obtained during the learning procedure. For instance, in \cite{yang2016joint}, such a behavior is reported. One possible reason for this to happen is that the cluster update step during the training doesn't go all the way till the end (see \emph{Number of iterations} in Section~\ref{sec:cupdates}).
\end{itemize}

\section{Related Methods}
\label{sec:related}
In this section we give a systematic overview of related work and analyze it within the framework of the proposed taxonomy. In Table~\ref{comparing} the methods are decomposed with regard to the following building blocks: Section~\ref{sec:arch} (network architecture), \ref{sec:layer} (used features), \ref{sec:ncloss} (non-clustering loss), \ref{sec:closs} (clustering loss), \ref{sec:comb} (loss combination), clustering algorithm, as well as their performance. Additional explanations to some of the methods are given in the following.

\newcommand{\centered}[2][c]{\begin{tabular}[#1]{@{}c@{}}#2\end{tabular}}

\begin{table*}[h!]

\resizebox{\textwidth}{!}{
\begin{tabular}{c|c|c|c|c|c|c|c|c|c|c}

METHOD & 
ARCH & 
\centered{FEATURES\\FOR\\CLUSTERING} & 
\centered{NON-\\CLUSTERING\\LOSS} & 
\centered{CLUSTERING\\LOSS} & 
\centered{COMBINING\\ THE LOSS\\ TERMS} & 
\centered{CLUSTERING\\ ALGORITHM} & 
\centered{NMI\\ MNIST} &
\centered{ACC\\ MNIST} &
\centered{NMI\\ COIL20} &
\centered{ACC\\ COIL20} \\ \hline \hline

\centered{JULE\\ \cite{yang2016joint}} &
CNN &
\centered{CNN\\output} &
- &
\centered{Agglomerative\\loss} &
- &
\centered{Agglomerative\\clustering} & 
0.915 &
- &
\textbf{1} &
- \\ \hline

\centered{CCNN\\ \cite{hsu2017cnn}} &
CNN  &
\centered{Internal\\CNN\\layer} &
- &
\centered{Cluster\\classification\\loss} & 
- & 
k-Means &
0.876 &
- & 
- &
-\\ \hline

\centered{SCCNN\\\cite{lukic2016speaker}} & 
\centered{CNN} &
\centered{CNN output} & 
\centered{Classification \\loss} & 
\centered{-} &
\centered{-} &
\centered{``Standard \\Clustering \\Method''} & 
- &
- &
- &
-
 \\ \hline

\centered{DEC\\ \cite{xie2016unsupervised}} &
MLP &
\centered{Encoder\\output} &
\centered{Autoencoder\\ reconstruction\\ loss} & \centered{Cluster\\assignment\\hardening} &
\centered{Pretraining\\ and\\ fine-tuning} &
\centered{Network \\estimates\\  centroids\\ (as trainable\\ parameters);\\ soft assignments\\ done based on\\ distance to\\ centroids}  &
0.800\footnotemark &
0.843 &
- &
- \\ \hline

\centered{DBC\\ \cite{li2017discriminatively}} &
CNN &
\centered{Encoder\\output} &
\centered{Autoencoder\\ reconstruction\\ loss} &
\centered{Cluster\\assignment\\hardening} &
\centered{Pretraining\\ and\\ fine-tuning} &
k-Means &
0.917 &
0.964 &
0.895 &
0.793 \\ \hline

\centered{DEPICT\\ \cite{dizaji2017deep}} &
CNN  &
\centered{Encoder\\output} &
\centered{Autoencoder\\ reconstruction\\ loss} &
\centered{Balanced-\\assignment} & 
\centered{Pretraining \\followed by\\ joint training}  &
\centered{Network \\estimates\\  centroids\\ (as trainable\\ parameters)\\ and predicts\\assignments} &
0.916 &
0.965 &
- &
-
\\ \hline

\centered{DCN\\ \cite{yang2016towards}} & 
MLP &
\centered{Encoder\\output} &
\centered{Autoencoder\\ reconstruction\\ loss} &
\centered{k-Means\\ loss} &
\centered{Alternating\\ between joint\\ training and\\ cluster updates} &
\centered{Network \\estimates\\  centroids;\\ soft assignments\\ based on distance} &
0.810 &
0.830 &
- &
- \\ \hline

\centered{DEN\\ \cite{huang2014deep}} &
MLP  &
\centered{Encoder\\output} &
\centered{Autoencoder\\ reconstruction\\ loss} &
\centered{Locality-\\preserving \\+ Group\\sparsity} & 
\centered{Joint\\training} &
k-Means &
- &
- & 
0.870 &
0.724
\\ \hline

\centered{Neural Clustering\\\cite{saito2017neural}} & 
MLP &
\centered{Concat.\\of encoder\\layers output} &
\centered{Autoencoder\\reconstruction\\loss} &
- &
- &
\centered{k-Nearest\\ Neighbors} & 
- & 
0.966 &
- &
- \\ \hline

\centered{NMMC\\\cite{chen2015deep}} & 
DBN &
\centered{Deepest \\DBN layer} &
\centered{Usual DBN loss} & 
\centered{Maximum \\ margin for mixture \\components} &
\centered{Pre-train DBN,\\ estimate number \\of clusters, \\fine-tune only the \\ deepest layers} &
- & 
- & 
- &
- &
-
\\ \hline

\centered{UMMC\\\cite{chen2017unsupervised}} & 
DBN &
\centered{Deepest \\DBN layer} &
\centered{Usual DBN loss} &
\centered{Locality-\\preserving \\loss + Cluster \\assignment \\hardening} &
\centered{Pretraining \\ followed by \\ joint training} &
\centered{k-Means} &
0.864 & 
- & 
0.891 &
-
\\ \hline

\centered{UGATC\\\citetext{Premachandran\\ \& Yuille 2016}} & 
\centered{GAN} &
\centered{Penultimate\\discriminator \\layer} &
\centered{Usual GAN loss} &
\centered{-} &
\centered{-} &
\centered{k-Means++}  & 
- &
- &
- &
-
\\ \hline

\centered{VaDE\\\cite{zheng2016variational}} & 
\centered{VAE} &
\centered{Encoder\\output} &
\multicolumn{2}{c|}{\centered{VAE loss with mixture model\\ for clusters}} & 
- &
\centered{Network \\estimates\\ centroids}  & 
- &
0.944 &
- &
-
 \\ \hline

\centered{TAGnet\\\cite{wang2016learning}} & 
\multicolumn{2}{c|}{\centered{Sparse-coding iterations \\time-unfolded  (reformulated) \\into a deep network}} & 
\centered{Sparse coding\\(implicit in \\architecture)} &
\centered{- Similarity graph\\regularization (implicit)\\- Cluster classification\\or maximum margin} &
\centered{Training on explicit\\loss; implicit losses\\ are part of the\\ architecture} &
\centered{Network predicts\\soft-assignments\\to clusters} &
0.651 &
0.692 &
0.927 &
\textbf{0.899} 
\\ \hline

 \centered{IMSAT\\\cite{hu2017learning}} & 
\centered{MLP} &
\centered{MLP output} &
\centered{Information \\maximization \\+ Self-\\Augmentation \\loss} &
\centered{-} &
\centered{Joint training} &
\centered{Network predicts \\soft-assignments \\to clusters}  & 
- &
\textbf{0.984} &
- &
-
 \\ \hline

Proposed Method& 
CNN &
\centered{Encoder\\ output} &
\centered{Autoencoder\\reconstruction\\loss} &
\centered{Cluster\\assignment\\hardening} &
\centered{Pretraining \\followed by\\ joint training} &
\centered{ Network \\estimates\\  centroids;\\ soft assignments\\ based on distance} &
\textbf{0.923} & 
0.961 &
0.853 &
0.781 \\

\end{tabular}
} 
\caption{Decomposition and comparison of existing methods based on the building blocks of the taxonomy. Quality values are taken from the respective original publications, unless stated otherwise. The method created as part of the case study outperforms other methods in NMI on MNIST and achieves balanced quality results across all metrics.}
\label{comparing} 
\end{table*}

\footnotetext{Values are taken from \citep{yang2016towards} since the DEC paper did not report NMI results.}

\textbf{Convolutional neural networks as feature extractors}\hspace{3pt}
Some methods utilize convolutional neural networks as feature extractors to improve clustering of image data. JULE \citep{yang2016joint} follows a hierarchical clustering approach and uses the agglomerative clustering loss as the sole loss function. 

In CCNN \citep{hsu2017cnn}, initialization of the k-means centroids is performed based on the softmax output layer of a convolutional neural network and the output of internal network layers are used as features. IMSAT \citep{hu2017learning} uses the information maximization loss and a self-augmentation loss. SCCNN \citep{lukic2016speaker} uses a convolutional network that was pretrained on a related, domain-specific task (speaker identification) as a feature extractor.

\textbf{Autoencoders as feature extractors}\hspace{3pt}
In contrast to the aforementioned methods, some of the most promising methods utilize autoencoders as feature extractors. Training is performed in two phases. In the first phase, the autoencoder is pretrained using a standard reconstruction loss. The second phase differs between the methods: In the second phase of DEC \citep{xie2016unsupervised}, the network is fine-tuned using the cluster assignment hardening loss.  
DEC is often used as a baseline for new publications. DBC \citep{li2017discriminatively} and DEPICT \cite{dizaji2017deep} are similar to DEC except for one aspect each: DBC utilizes a convolutional autoencoder in order to improve clustering of image datasets and  DEPICT adds a balanced assignments loss to the clustering loss to alleviate the danger of obtaining degenerate solutions. 

In contrast to DEC and similar methods, DCN \citep{yang2016towards} jointly trains the network with both the autoencoder reconstruction loss and the $k$\nobreakdash-means clustering loss in the second phase. 
In, DEN \cite{huang2014deep}, the second phase is comprised of joint training with a combination of the reconstruction loss, a locality-preserving loss, and a group sparsity loss. In Neural Clustering \cite{saito2017neural}, the second training phase does not exist and no additional clustering loss is used. The good results can possibly be attributed to the fact that multiple network layers are concatenated and used as the clustering features.
 
\textbf{Other deep architectures as feature extractors}\hspace{3pt}
NMMC \citep{chen2015deep} and UMMC \citep{chen2017unsupervised} each use a deep belief network for feature extraction. UGTAC \citep{premachandran2016unsupervised} uses the penultimate layer of the discriminator in the DCGAN \citep{radford2015unsupervised} architecture as features for k-means++ clustering. VaDE \citep{zheng2016variational} uses a variational autoencoder in combination with a mixture of Gaussians.

\textbf{Other methods}\hspace{3pt}
Instead of directly using a neural network to extract features, infinite ensemble clustering \citep{liu2016infinite} uses neural networks to generate infinite ensemble partitions and to fuse them into a consensus partition to obtain the final clustering.

\begin{figure}[h]
  \includegraphics[width=\linewidth]{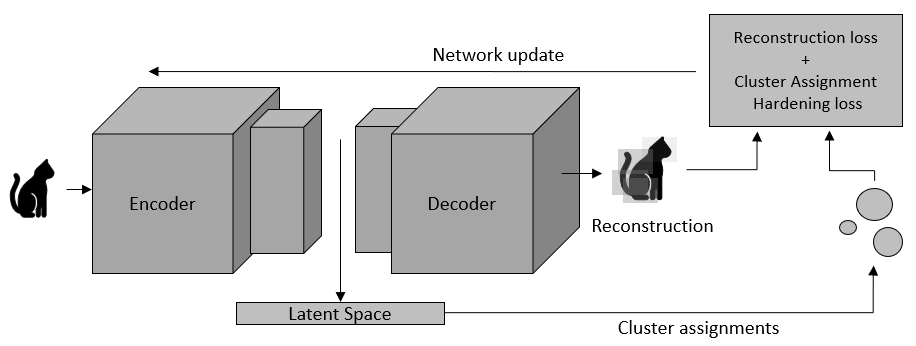}
  \caption{Our proposed method is based on a fully convolutional autoencoder trained with reconstruction and cluster hardening loss as described in Section \ref{sec:ncloss} and \ref{sec:closs}. It results in clustering-friendly feature space with no risk of collapsing.}
  \label{fig:proposed}
\end{figure}

\section{Case Study: New Method}
\label{sec:proposed}

We argue that the proposed taxonomy (Section~\ref{sec:taxonomy}) enables a more systematic and analytical approach to creating new methods. By decomposing existing methods according to building blocks of the taxonomy, as shown in Table~\ref{comparing}, researchers and practitioners can analyze commonalities amongst good methods and attribute limitations to specific building blocks. To show that the taxonomy makes this approach easier, we conduct a case study, in which we create a method that overcomes the limitations of the previous approaches and achieves better results for the task at hand.

A high-level overview of our method is provided in Figure~\ref{fig:proposed}. 
We will make the code publicly available.

Since the dataset that is to be clustered consists of images, we chose a convolutional architecture for the architecture building block (Section \ref{sec:arch}) and an autoencoder reconstruction loss as the non-clustering loss building block (Section \ref{sec:ncloss}).

After pretraining the network with this configuration, we chose to keep the non-clustering autoencoder reconstruction loss and add the cluster hardening loss as the clustering loss building block (Section \ref{sec:closs}).
We decided to keep the non-clustering loss in this phase, because other methods (e.g.~DEC and DBC) which omit it, have a theoretical risk of collapsing clusters and have also empirically shown to yield worse clustering quality. This is because valuable feature detectors that were previously learned during pretraining can be destroyed. Keeping the non-clustering loss helps to keep those detectors stable. For a similar reason, we decided to avoid a training procedure with alternating losses as for instance used by DBC, DEN, and UMMC. The necessity for this alternation originates in the hard assignments as enforced by the $k$\nobreakdash-means loss, which is why we chose soft cluster assignments and cluster hardening loss instead. This loss is achieved via introducing a clustering layer, which aims at estimating the cluster assignments based on equation~\eqref{eq:estimate}.

Once both training phases (pretraining on non-clustering loss and fine-tuning on both losses) are completed, the network's encoder has learned to map its input to a clustering-friendly space. Additionally, the resulting network is capable of estimating the cluster assignments. However, based on the previous success of re-running clustering from scratch after training (see Section~\ref{sec:aftertrain}), we run the $k$\nobreakdash-means algorithm on the network's output representation.

\begin{figure*}
    \centering
    \captionsetup{justification=centering}
	\begin{subfigure}{\textwidth}
		\begin{subfigure}[b]{0.32\textwidth}
			\includegraphics[width=\textwidth,height=0.16\textheight]{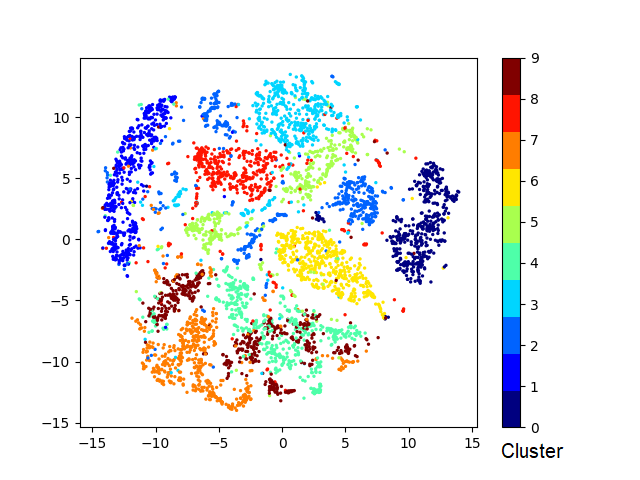}
			\caption{k-Means}
			\label{fig:mnist-km}
		\end{subfigure}
		\begin{subfigure}[b]{0.32\textwidth}
		    \includegraphics[width=\textwidth,height=0.16\textheight]{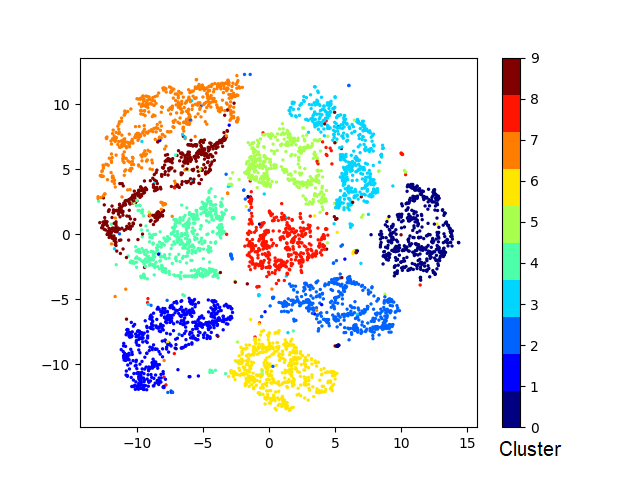}
			\caption{Autoencoder + k-Means}
			\label{fig:mnist-ae}
		\end{subfigure}
		\begin{subfigure}[b]{0.32\textwidth}
		    \includegraphics[width=\textwidth,height=0.16\textheight]{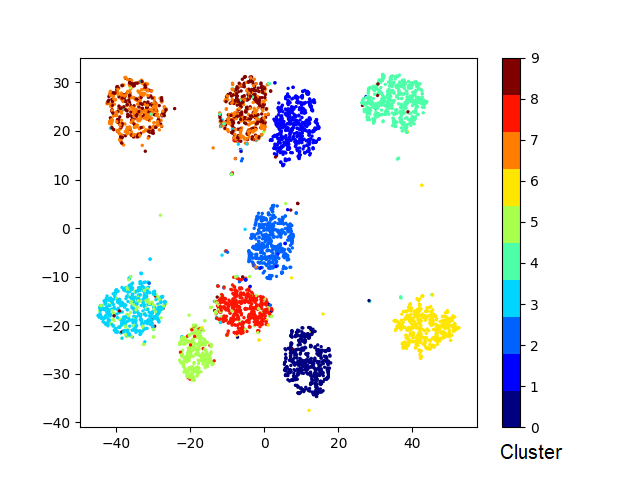}
			\caption{Proposed} 
			\label{fig:mnist-p}
		\end{subfigure}
	\end{subfigure}
	\caption{$t$\nobreakdash-SNE visualizations for clustering on MNIST dataset in (a) Original pixel space, (b) Autoencoder hidden layer space and (c) Autoencoder hidden layer space with the proposed method.}
	\label{fig:results-tsne-mnist}
\end{figure*}
\begin{figure*}
    \centering
    \captionsetup{justification=centering}
	\begin{subfigure}{\textwidth}
		\begin{subfigure}[b]{0.32\textwidth}
			\includegraphics[width=\textwidth,height=0.16\textheight]{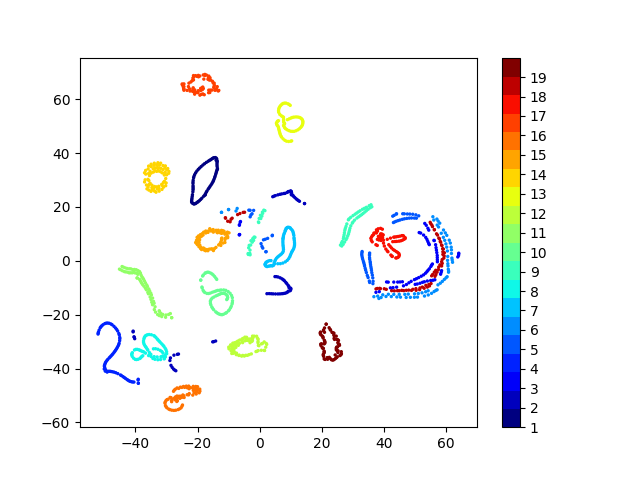}
			\caption{k-Means}
			\label{fig:coil-km}
		\end{subfigure}
		\begin{subfigure}[b]{0.32\textwidth}
		    \includegraphics[width=\textwidth,height=0.16\textheight]{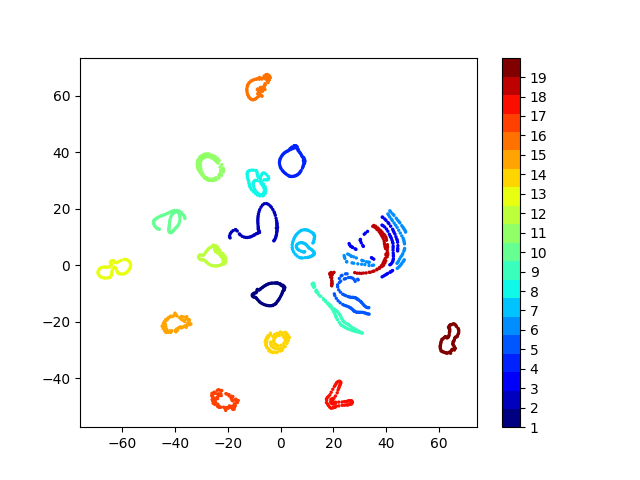}
			\caption{Autoencoder + k-Means}
			\label{fig:coil-ae}
		\end{subfigure}
		\begin{subfigure}[b]{0.32\textwidth}
		    \includegraphics[width=\textwidth,height=0.16\textheight]{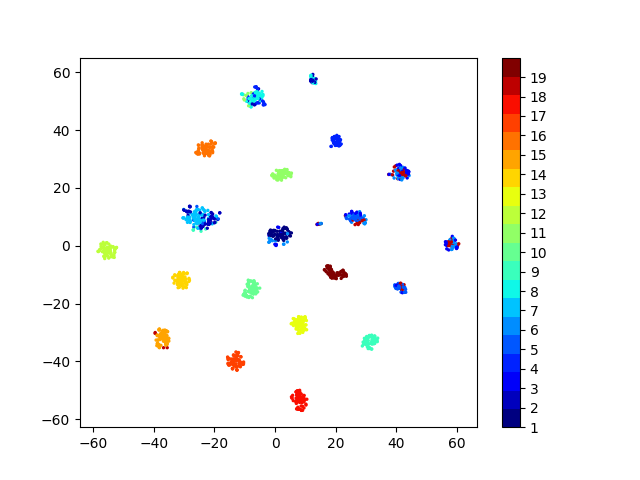}
			\caption{Proposed} 
			\label{fig:coil-p}
		\end{subfigure}
	\end{subfigure}
	\caption{$t$\nobreakdash-SNE visualizations for clustering on COIL20 dataset in (a) Original pixel space, (b) Autoencoder hidden layer space and (c) Autoencoder hidden layer space with the proposed method.}
	\label{fig:results-tsne-coil}
\end{figure*}

\section{Experimental Results}
\label{sec:results}
In this section we evaluate our proposed method and compare it to the methods previously introduced in Section \ref{sec:related}.

\textbf{Validation Metrics}\hspace{3pt} As evaluation metric we use clustering accuracy (ACC) and normalized mutual information (NMI) \citep{strehl2002cluster,vinh2010information,cai2011locally}. Both metrics have a range of $[0,1]$ with $1$ being the perfect clustering. Experiments are performed on datasets for which meaningful cluster assignments are known (hence ACC and NMI can be computed) but these assignments are not provided to the methods.

\textbf{Experimental Setup}\hspace{3pt} A structured hyper-parameter search led us to training our model with a learning rate of $0.01$ and a momentum of $0.9$. We used batch normalization \citep{ioffe2015batch} and L2 regularization.

\textbf{Datasets}\hspace{3pt} The experiments were performed on two publicly available datasets:

\begin{itemize}
    \item \textbf{MNIST}: Consists of $70000$ images of hand-written digits of $28\times28$ pixel size. The digits are centered and size is normalized \citep{lecun1998mnist}.
    \item \textbf{COIL20}: Contains $1440$, $32 \times 32$ gray scale images of $20$ objects. For each object, $72$ images were taken with a 5 degrees distance \citep{nene1996columbia}.
\end{itemize}

\textbf{Performance}\hspace{3pt} In Table~\ref{comparing}, the clustering performance is denoted with respect to the two metrics and two datasets. The results for the other methods are taken from the respective publications. In comparison to those other methods, our method is able to surpasses the state of the art for NMI on the MNIST dataset with a value of 0.923 and performs comparably on others. It should also be noted, that compared to other state of the art methods, our results are more balanced with respect to the datasets. For instance, while TAGNet \citep{wang2016learning} defines the state of the art on the COIL20 dataset, it performs quite bad on MNIST.

Our results show that the taxonomy enables the creation of high-performing methods in a structured and analytical way. Researches are now able to selectively recombine or replace distinct aspects of existing methods instead of relying on a experiment and discovery based approach.

Figure~\ref{fig:results-tsne-mnist} and \ref{fig:results-tsne-coil} visualize the clustering spaces at different training stages of the proposed network. The clustering spaces are $120$-dimensional and $320$-dimensional for MNIST and COIL20, respectively. The true cluster labels are indicated by the different colors. The visualizations show that the proposed method results in a more clustering-friendly space compared to the original image space and the space before jointly training with the clustering loss.

\section{Conclusion}
\label{sec:conclusion}
In this paper, we proposed a universal taxonomy for clustering methods that utilize deep neural networks. The taxonomy was created by decomposing existing methods and analyzing the building blocks for properties that are indicative of desirable or undesirable results. Our case study shows that by using this taxonomy, researches and practitioners can derive new clustering methods in a systematic and analytical way to suit their task. The method we constructed as part of the case study was able to outperform the previous state of the art method on the MNIST dataset. Following the same approach, the proposed taxonomy can thus inspire a myriad of new methods.

\bibliographystyle{apalike}
\bibliography{main_bib}

\end{document}